\documentclass[runningheads]{llncs}

\usepackage{adjustbox}
\usepackage{graphicx}
\usepackage[caption=false]{subfig}
\usepackage{svg}
\usepackage{hyperref}
\hypersetup{
    colorlinks=true,
    linkcolor=blue,
    filecolor=magenta,      
    urlcolor=cyan
}
\usepackage[firstpage]{draftwatermark}

\SetWatermarkLightness{0.9} 
\SetWatermarkFontSize{10cm} 
\SetWatermarkScale{5} 
\SetWatermarkText{PREPRINT}

\usepackage{tikz}
\usepackage{url}
\usetikzlibrary{decorations.pathreplacing,positioning,shapes}
\usepackage{tabularx}
\usepackage{multirow}

\usepackage{media9}
\usetikzlibrary{calc}
\usetikzlibrary{hobby}
\usetikzlibrary{arrows.meta}
\usepackage{amsfonts,bm,amsmath}

\usepackage{graphicx}

\usepackage{color}

\usepackage{array}
\newcolumntype{P}[1]{>{\centering\arraybackslash}p{#1}}

\begin{document}
\title{Strengthening structural baselines for graph classification using Local Topological Profile}

\titlerunning{Graph classification using Local Topological Profile}

\author{Jakub Adamczyk\orcidID{0000-0003-4336-4288} \and
Wojciech Czech\orcidID{0000-0002-1903-8098}}

\authorrunning{J. Adamczyk, W. Czech}

\institute{AGH University of Science and Technology, Kraków, Poland \\
\email{\{jadamczy,czech\}@agh.edu.pl}}

\maketitle

\begin{abstract}

We present the analysis of the topological graph descriptor Local Degree Profile (LDP), which forms a widely used structural baseline for graph classification. Our study focuses on model evaluation in the context of the recently developed fair evaluation framework, which defines rigorous routines for model selection and evaluation for graph classification, ensuring reproducibility and comparability of the results. Based on the obtained insights, we propose a new baseline algorithm called Local Topological Profile (LTP), which extends LDP by using additional centrality measures and local vertex descriptors. The new approach 
provides the results outperforming or very close to the latest GNNs for all datasets used. Specifically, state-of-the-art results were obtained for 4 out of 9 benchmark datasets. We also consider computational aspects of LDP-based feature extraction and model construction to propose practical improvements affecting execution speed and scalability. This allows for handling modern, large datasets and extends the portfolio of benchmarks used in graph representation learning. As the outcome of our work, we obtained LTP as a simple to understand, fast and scalable, still robust baseline, capable of outcompeting modern graph classification models such as Graph Isomorphism Network (GIN). We provide open-source implementation at \href{https://github.com/j-adamczyk/LTP}{GitHub}.
 
\end{abstract}

\section{Introduction}
 
Graph classification is an essential variant of supervised learning problems, gaining popularity in many scientific fields due to the growing volume of structured datasets, which encode pairwise relations between modeled objects of different types. The applications of graph classification algorithms range from cheminformatics \cite{hu2020open}, where high-level properties of molecules such as toxicity or mutagenicity are predicted, to sociometry \cite{yanardag2015deep}, biology \cite{zhang2021graph} and technology \cite{li2019graph}, tackling different classes of complex networks, whose non-trivial dynamics can be explained by learning structural patterns. 

Graph classification poses the inherent problem of measuring the dissimilarity between objects which do not lie in metric space but have combinatorial nature. This challenge is typically addressed by extracting isomorphism-invariant representations in the form of feature vectors \cite{zhang2021systematic} (also called graph embeddings, descriptors, fingerprints) or by constructing explicit pairwise similarity measures known as graph kernels \cite{kriege2020survey}. More recently, the graph embedding problem was successfully reformulated within the framework of deep convolutional neural networks. Adopting the concept of convolution to vertex neighborhoods by introducing hierarchical iterative operators on multidimensional states of vertices allowed for building task-specific, low-dimensional representations for vertices, edges, and, after global pooling, the whole graph \cite{wu2020comprehensive}. 

Baselines are the crucial elements of the fair comparison frameworks used in machine learning. As deep learning methods become increasingly powerful, the baseline algorithms used for their evaluation should also provide competitive results, forming good reference points for analyzing algorithms' performance. The recent development of Graph Neural Networks (GNNs), which automatically extract task-relevant features via deep learning, increased the number of attempts to solve various graph classification tasks \cite{zhang2018end,zhou2022graph}. Nevertheless, fair evaluation practices were frequently neglected in reported studies, and only recently the need for more rigorous model evaluation was highlighted \cite{errica2019fair}. This increased the demand for more powerful yet simple and fast baseline methods.

Motivated by recent findings regarding the discriminative power of Local Degree Profile (LDP) \cite{cai2018simple}, which, together with SVM as the classification model, were proven to be competitive with the newest GNN models, we study the robustness and scalability of the new Local Topological Descriptor (LTP) built using histograms of specific descriptors representing vertex and edge structural features. We also use Random Forest classifier instead of SVM to reduce the sensitivity of baseline to hyperparameter tuning \cite{probst2019tunability}. All experiments are performed in the regime of fair evaluation framework \cite{errica2019fair} to ensure replicability and to correct inaccuracies present in some earlier works (such as reporting accuracy on validation set). We also propose performance improvements in the implementation of LDP and LTP, resulting in better scalability and enabling the computation on large and dense social network benchmarks.

The key contributions of this work are the thorough analysis of the graph classification baseline composed of LDP and SVM, reporting limitations of this approach, the proposal of a new topological baseline utilizing Random Forest and experimental evaluation showing its robustness compared to the state-of-the-art. In addition, we present a modular software framework for LDP/LTP-based graph classification together with the associated open-source Python code shared on \href{https://github.com/j-adamczyk/LTP}{GitHub}.

\section{Related works}

In the early works related to graph comparison, the concept of graph edit distance (GED) was introduced \cite{bunke1997relation}. It was based on calculating the optimal sequence of elementary operations (adding/removing vertex/edge) required to transform one graph into another. 
Computational complexity prevented GED algorithms from being widely used for larger graphs ($> 1000$ vertices). Nevertheless, multiple successful attempts at classifying attributed graphs were reported based on benchmark dataset \cite{riesen2008iam}. The IAM graph database \cite{riesen2008iam} formed the first consistent framework for comparing the efficiency of graph classification algorithms based on GED or graph embedding methods. 

Graph embedding forms the comprehensive field of methods and applications aimed at the generation of multidimensional graph invariants/descriptors, which can be recognized as graph feature extraction or feature engineering \cite{zhang2021systematic}. Graph descriptors can be assigned to the vertex, the edge, or the graph itself. Representing a graph as a vector enables using a multitude of unsupervised and supervised machine learning algorithms suitable for tabular data. The most popular graph invariants come from the field of complex networks and spectral graph theory. They are represented by several graph centrality measures, \textit{clustering coefficient}, \textit{efficiency} and permutation invariants constructed from the eigenpairs of Laplace matrix \cite{wilson2005pattern}. Generic-purpose vertex and edge descriptors can be aggregated to form a high-dimensional graph representation such as B-matrix \cite{czech2012invariants} or, after including vertex attributes, even more expressive relation order histograms \cite{lazarz2021relation}. Graph descriptors can be also extracted by mining frequent patterns/subgraphs, resulting in the topological fingerprint suitable for structural pattern recognition but also querying graph databases \cite{kuramochi2004efficient}. This approach was further extended by introducing domain-specific representations such as molecular fingerprints \cite{rogers2010extended}, which are widely used in the prediction of biochemical properties. As graph embedding algorithm can be adjusted to the domain, graph type (e.g. attributed vs. non-attributed), or even available computing resources, the topological descriptors still represent promising are for graph feature engineering and, as presented in \cite{cai2018simple}, can compete with state-of-the-art graph representation learning techniques.

The concept of graph substructure mining was generalized in the form of graph kernels \cite{kriege2020survey}, which assess the structural similarity between two graphs by pairwise comparison of their subcomponents. Most typically, the concept of R-convolution \cite{haussler1999convolution} is applied as a generic purpose convolution framework for discrete structures. 
One of the most interesting representatives of this group are Weisfeiler-Lehman kernels designed for subtrees, edges, shortest paths, and whole graphs \cite{shervashidze2011weisfeiler}. They utilize the concept of Weisfeiler-Lehman test of isomorphism, which was also used in the construction of Graph Isomorphism Networks (GIN). Another family of graph kernels uses the concept of optimal assignment \cite{frohlich2005optimal} to reduce the number of pairwise sub-kernel computations required to obtain a similarity value. 
In this work, we focus on explicit graph embedding and graph representation learning, skipping graph kernels as a less feasible solution for scalable graph classification.

In case of graph embedding and graph kernels, the domain-specific knowledge can be incorporated by designing specific substructure descriptors with the help of experts. Such an approach can be treated as an example of feature engineering. The different method, providing automatic, task-specific graph feature extraction, is represented by Graph Representation Learning models exemplified by Graph Neural Networks (GNNs). They form a modern and extensively studied framework for graph classification, with dozens of available models and specific taxonomy \cite{liu2022taxonomy}. Graph Isomorphism Networks (GIN) \cite{xu2018powerful} were designed to be as powerful as Weisfeiler-Lehman isomorphism test in discriminating graphs. They were reported to achieve state-of-the-art results on graph classification benchmarks; therefore, they will be used as the main reference point for evaluating our method. GraphSAGE \cite{hamilton2017inductive} is a general inductive framework for different convolutional GNNs, providing a new neighborhood sampling method, which ensures fixed-size aggregation sets to limit computational overhead related to processing hubs, present, e.g., in social networks. The aggregation function for node states can be treated as a hyperparameter and tuned on the validation set. The newer DiffPool model \cite{ying2018hierarchical} generates hierarchical graph representations using a differentiable graph pooling layer. It assigns nodes to clusters to achieve coarse-graining of input for the next layer, which reduces computation time. Edge-Conditioned Convolution (ECC) model \cite{simonovsky2017dynamic} introduces convolutions over local graph neighborhoods using edge labels and custom coarsening procedure subsampling vertices on pooling layers to reduce graph size. The high classification accuracy was achieved by ECC on molecular datasets. Also, Deep Graph Convolutional Neural Network (DGCNN) model \cite{zhang2018end} proposes custom localized graph convolution similar to spectral filters and related to Weisfeiler-Lehman subtree kernel. Additionally, the new SortPooling layer is introduced, enabling standard neural network training on graphs. All models mentioned in this paragraph will be used in the evaluation of the new LTP baseline.

\section{Methods}

Graph classification tasks can be organized into well-defined pipeline. First, the graph is typically represented as a sparse adjacency matrix. Optionally, vertex and edge feature matrices can be used, if they are available. The matrices provide the input to the feature extraction algorithm, which outputs a feature vector representing a graph embedding in a metric space. Next, a tabular classification algorithm is used. For explicit feature extraction methods, such as LDP or graph kernels, the graph invariants are calculated by an algorithm distinct from the classifier. This allows using arbitrary algorithms for both parts. For graph representation learning methods, such as GNNs, those representations are typically learned end-to-end using a differentiable framework and gradient-based optimization, with multilayer perceptron (MLP) as a classifier. This potentially increases flexibility and makes embeddings more task-related, but requires vastly more data and computational resources.

\textbf{Local Degree Profile (LDP)} \cite{cai2018simple} proposes a feature extraction based on vertex degree statistics, which are calculated for each node in the graph and then aggregated into the embedding vector. Following conventions from \cite{cai2018simple}, we denote the graph as $G(V,E)$, where $V$ is the set of vertices (or nodes) $v$ and $E$ is the set of edges $e$. Degrees of neighboring nodes form a multiset $DN(v) = \{ \text{degree}(v) | (u, v) \in E\}$. For each node, we then calculate the following statistics: $\text{degree}(v)$, $\min(DN)$, $\max(DN)$, $\text{mean}(DN)$, $\text{std}(DN)$. This way, for each node, we obtain the summary statistics of itself and its 1-hop neighborhood. They are then aggregated for the whole graph by calculating a histogram or empirical distribution function (EDF) over each feature. They are concatenated for all features, forming a final graph embedding. The number of bins used for aggregation and the choice between histogram and EDF are hyperparameters. There are also additional hyperparameters reflecting the method of preprocessing the features before the aggregation. Normalization can be applied: separately per graph, dividing the degrees by the highest value (this results in representing the feature value relative to the rest of the graph), or for the whole dataset, dividing by the highest degree in the dataset. Also, based on the observation that node degrees follow a power law for social networks, one can use log scale for aggregating features.

Any additional node- or edge-based structural descriptors can be included in the LDP in the form of histograms. The authors experiment with multiple ones: neighbors degree sum, lengths of shortest paths, closeness centrality, Fiedler vector and Ricci curvature. They remark that only shortest paths gave visible advantage, but could only be calculated in reasonable time on bioinformatics datasets, which have small molecular graphs. However, the gains using the shortest paths are consistent, about 2\%, indicating that incorporating edge-based information can be beneficial. It should be noted that additional descriptors rapidly increase the dimensionality of the resulting embedding, which may result in degraded performance due to the curse of dimensionality, so only a limited number of well-chosen descriptors should be used.

Over the years, a vast number of node and edge descriptors were proposed. Among them, there are three commonly used groups, describing very different structural properties of the graph: centrality measures, link prediction indexes, and sparsification scores. Centrality indicates the importance or the influence of the node in the graph. Different measures focus on, e.g., how much information flows though a given node, or how many walks go through that node. They can also be defined for edges, indicating importance of connections in the graph. Link prediction indexes describe edges and are used to suggest the new edges to be added to the graph. They analyze the neighborhoods of the nodes, assigning high scores to the potential edges between nodes that share a large portion of their neighbors, or which have neighborhoods leading to shorter paths between them. Graph sparsification algorithms aim to eliminate edges, which are the least important for keeping the overall structure of the graph, especially in relation to hub nodes and communities. They aim to locally incorporate more global information about graph topology, assigning higher scores to more important edges.

During preliminary experiments, we surveyed descriptors representing each of those groups and available in the Networkit \cite{networkit} library. While almost all gave promising results, we selected one from each group: edge betweenness centrality, Jaccard Index and Local Degree Score. The selection was based on intuitions that those particular descriptors will bring more edge-focused or more global information than only node degree statistics, enhancing the discriminative power of the baseline.

\textbf{Edge betweenness centrality (EBC)} 
\cite{girvan2002community} 
is a centrality measure based on shortest paths, which measures how much influence the edge has over the flow of information in the network. It is defined as the fraction of the shortest paths in the graph going through the edge $e = (u,v)$:
\begin{equation}
EBC(e) = \sum_{s, t; (s, t) \neq (u, v)} \frac{\sigma_{st}(e)}{\sigma_{st}},
\end{equation}
\noindent
where $\sigma_{st}$ is the total number of shortest paths between nodes $s$ and $t$, and $\sigma_{st}(e)$ is the number of those paths that go through $e$. This can be computed using Floyd-Warshall algorithm, which will give infinity values for disconnected graphs; we simply omit them in our implementation. We selected this descriptor, since it is based on shortest paths, which gave the good results in \cite{cai2018simple}, and also takes into consideration the cyclic structure of the graph, e.g., it distinguishes molecules with linear scaffolds vs those with more ring-like topology.

\textbf{Jaccard Index (JI)}
\cite{jaccard1912distribution} is a normalized overlap between node neighborhoods $N(u)$ and $N(v)$:
\begin{equation}
JI(u, v) = \frac{|N(u) \cap N(v)|}{|N(u) \cup N(v)|}
\end{equation}
\noindent
We calculate it for the existing edges $e=(u,v)$ in the graph, obtaining a descriptor of a 3-hop subgraph. This feature should better discriminate between graph with visible community substructures and those without such node clusters.

\textbf{Local Degree Score (LDS)} \cite{lindner2015structure} was proposed to detect edges between hubs, i.e. nodes with locally high degree, and keep only those edges after graph sparsification. For each node $v$, the rank of its neighbor $u$, $\text{rank}(v, u)$ is the number of neighbors of $v$ with degree lower than $u$. Note that this is asymmetrical, i.e. $\text{rank}(u, v) \neq \text{rank}(v, u)$. For each edge, the Local Degree Score is defined as:
\begin{equation}
LDS(e) = \max\left(
1 - \frac{\ln\text{rank}(v, u)}{\ln\text{degree}(v)},
1 - \frac{\ln\text{rank}(u, v)}{\ln\text{degree}(u)}
\right),
\end{equation}
\noindent
which is simply taking the higher value from perspective of $u$ or $v$, since either of them can be the hub node, giving a high LDS value. We selected this feature, since it can indicate the dispersion of nodes in the graph. If there are few edges with high LDS, it indicates that there are a few well separated clusters in the graph, centered around hub nodes, e.g., communities or well-connected functional groups in chemistry.

We propose to use the LDP descriptors together with the additional features described above, creating the \textbf{Local Topological Profile}. This method incorporates additional graph topology information in a local fashion, enhancing LDP with more discriminative power. Of course, this increases the computational cost, which we discuss below.

LDP authors remark that shortest path lengths are not used for social network datasets due to unreasonably long computing time. Their implementation, however, uses NetworkX \cite{networkx} for computing graph descriptors, which is a Python library, performing sequential computations. Instead, we propose to use Networkit \cite{networkit}, a parallelized C++ library. This way, we are able to utilize modern CPUs with multiple cores and compute descriptors in parallel. We do not perform a timing comparison, as we also could not finish computation with NetworkX in any reasonable time. The experiments on subsets of datasets indicate that Networkit is at least a one or two degrees of magnitude faster even on much smaller, molecular graphs. For this reason, our whole implementation of descriptors computation is based on Networkit.

Classification algorithm applied to feature vectors has a direct influence on both accuracy and scalability. LDP used Support Vector Machine (SVM) with a Gaussian kernel, which is a powerful classifier traditionally used with graph kernels, since they work well with small datasets. However, they are not scalable, since kernel calculation alone takes $O(n^2)$ for $n$ graphs in the dataset. Moreover, they are sensitive to hyperparameter choice \cite{probst2019tunability}, hence requiring extensive tuning to obtain good results. In addition, they are typically trained with the Sequential Minimal Optimization (SMO) algorithm, which is inherently sequential, not utilizing modern CPUs with multiple cores. Linear SVMs, while faster to compute, typically give worse results, which the authors of LDP also observe.

We propose to change SVM to a Random Forest (RF) classifier. It is bagging ensemble of decision trees, which means that each tree can be trained independently in parallel, increasing scalability. Decision tree induction is also very fast, relying on a greedy top-down algorithm. They typically give good results with default hyperparameters, requiring only a sufficiently high number of trees \cite{probst2019randomforest}. In preliminary study, we did not observe any significant effect of hyperparameter tuning, even with large hyperparameter grids, hence we skipped this step.

More importantly, in contrast to LDP \cite{cai2018simple}, we use a fair evaluation protocol with test sets. We use the fair comparison procedure from \cite{errica2019fair}, adapted in the following way. The datasets and their statistics are summarized in Table \ref{table_datasets}. We use the same test splits, and apply 10-fold CV for testing. However, since our baseline is fast to compute, we can afford to perform inner 5-fold CV for validation and hyperparameter tuning, instead of holdout. Following \cite{errica2019fair}, we report mean and standard deviation of accuracy on test sets.

\begin{table}
\caption{Statistics of datasets used, following \cite{errica2019fair}.}
\label{table_datasets}
\scriptsize
\centering
\begin{tabular}{|l|c|c|c|c|}
\hline
\textbf{Dataset} & \textbf{\# Graphs} & \textbf{Avg. \# Nodes} & \textbf{Avg. \# Edges} & \textbf{\# Classes} \\ \hline
DD               & 1178               & 284.32                 & 715.66                 & 2                   \\ \hline
NCI1             & 4110               & 29.87                  & 32.30                  & 2                   \\ \hline
PROTEINS         & 1113               & 39.06                  & 72.82                  & 2                   \\ \hline
ENZYMES          & 600                & 32.63                  & 64.14                  & 6                   \\ \hline
IMDB-B           & 1000               & 19.77                  & 96.53                  & 2                   \\ \hline
IMDB-M           & 1500               & 13.00                  & 65.94                  & 3                   \\ \hline
REDDIT-B         & 2000               & 429.63                 & 497.75                 & 2                   \\ \hline
REDDIT-5K        & 4999               & 508.82                 & 594.87                 & 5                   \\ \hline
COLLAB           & 5000               & 74.49                  & 2457.78                & 3                   \\ \hline
\end{tabular}
\end{table}

\noindent We tune the following hyperparameters for LDP (the same as the authors of \cite{cai2018simple}):
\begin{itemize}
    \item number of bins: [30, 50, 70, 100]
    \item aggregation: [histogram, EDF]
    \item normalization: [none, graph, dataset]
    \item log scale: [false, true]
\end{itemize}

\noindent We tune the following hyperparameters for SVM (the same as authors of \cite{cai2018simple}):
\begin{itemize}
    \item C (regularization): [$10^{-3}$, $10^{-2}$, ..., $10^2$, $10^3$]
    \item $\gamma$ (Gaussian kernel bandwidth): [$10^{-2}$, $10^{-1}$, ..., $10^1$, $10^2$]
\end{itemize}

\noindent For RF, we do not perform tuning, instead setting the following parameters (based on Scikit-learn defaults) for dataset with $n$ samples and $d$ features: 500 trees, minimizing \textit{Gini impurity}, using $\sqrt{d}$ features, sampling $n$ samples with replacement.

We use PyTorch Geometric \cite{Fey2019pytorchgeometric} for data loading and computing node degree features, Networkit \cite{networkit} for computing EBC, JI and LDS descriptors, and Scikit-learn to implement SVM and RF. We perform all experiments using 12th Gen Intel Core i7-12700KF 3.61 GHz processor with 32 GB of RAM. Feature extraction processes graphs sequentially, while feature calculation is done in parallel, using all available cores. We use all available cores for RF (\verb|n_jobs=-1|) and for grid search. We performed experiments to answer the following questions:
\begin{enumerate}
    \item Can we improve training speed and prediction accuracy, using RF instead of kernel SVM? If so, by how much?
    \item Is tuning all LDP hyperparameters necessary? Can we eliminate some hyperparameters, or set reasonable defaults, in order to decrease tuning time?
    \item Do additional descriptors increase prediction accuracy? Can we use all 3 additional descriptors to get the best average improvement?
    \item What is the difference in training speed between the original LDP (using SVM and with hyperparameter tuning) and our proposed LTP (using RF and without tuning)?
    \item How does LTP compare against baselines from \cite{errica2019fair} and GNNs?
\end{enumerate}

\section{Results and discussion}
\label{section_results}

\begin{figure}[htb]
\begin{center}
\includegraphics[width=0.8\textwidth]{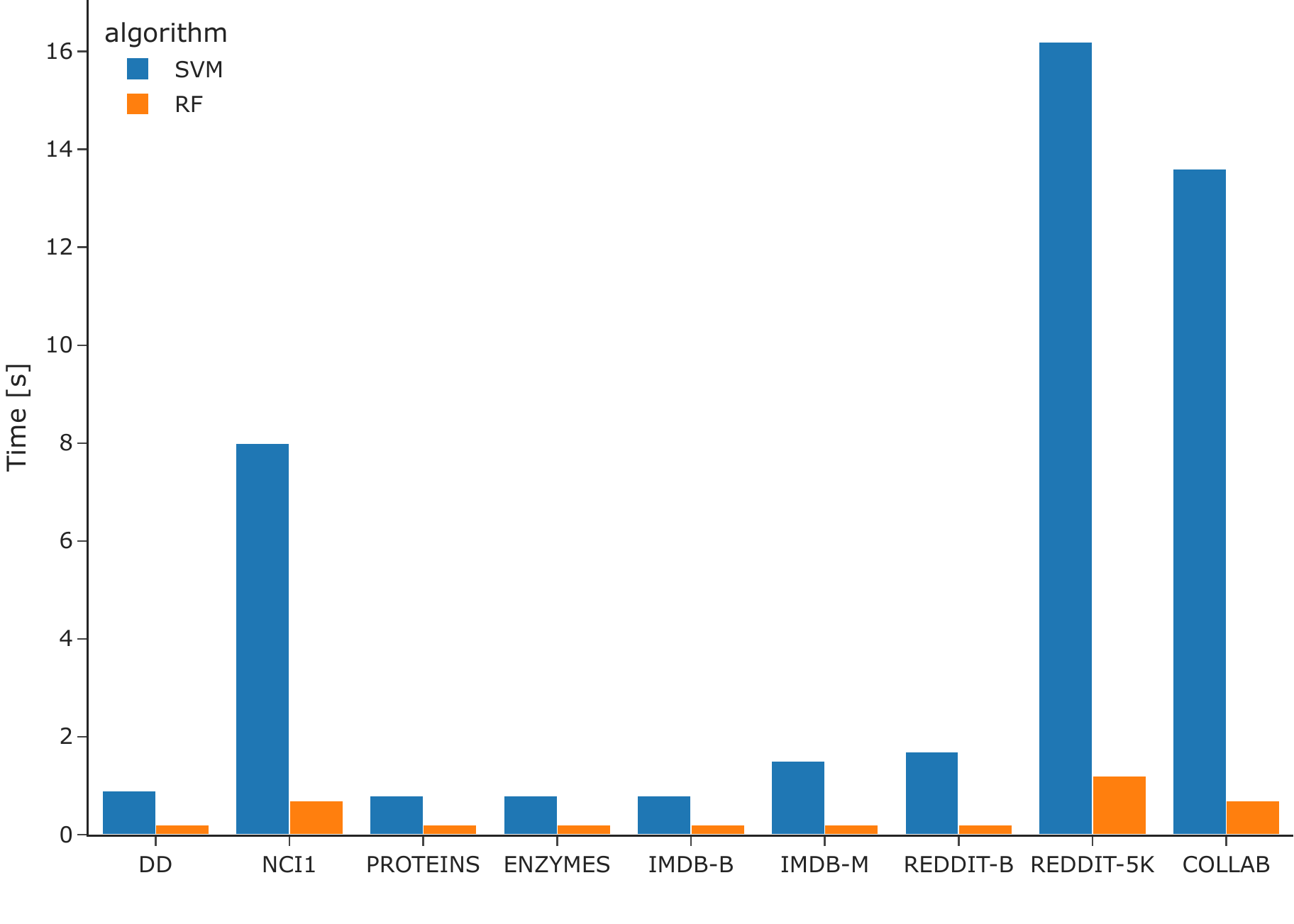}
\caption{Training time using LDP features: SVM vs RF clasifier.}
\label{fig:svm_vs_rf}
\vspace{-0.5cm}
\end{center}
\end{figure}

The first experiment concerned comparison of LDP prediction accuracy when using RF (without tuning) instead of SVM (with tuning). We include both linear and kernel SVM results. We used a reasonable default values based on LDP paper \cite{cai2018simple}: 50 bins, histogram aggregation, normalization per graph, and linear scale. As shown in Table \ref{table_svm_vs_rf_v2}, RF always gave better results than both linear and kernel SVM. Interestingly, in some cases linear SVM outperformed kernel SVM, contradicting findings in \cite{cai2018simple}. The average improvement of RF over SVMs across all datasets is 3.4\%, but can be as high as 7.3\% on ENZYMES or 5.9\% on DD. Additionally, the timings presented in Figure \ref{fig:svm_vs_rf} indicate that RF is about an order of magnitude faster than SVM, being the result of both a more parallelizable algorithm and no need for hyperparameter tuning. Based on this finding, we only use RF in further experiments.

\begin{table}[htb]
\caption{Classification accuracy on testing sets using LDP features and three analyzed models. The best result for each dataset is marked in bold.}
\label{table_svm_vs_rf_v2}
\scriptsize
\makebox[\textwidth][c]{
\begin{tabular}
{|p{0.18\textwidth}|P{0.16\textwidth}|P{0.16\textwidth}|P{0.16\textwidth}|}
\hline
\textbf{Dataset} & \textbf{Linear SVM} & \textbf{Kernel SVM} & \textbf{RF} \\
\hline
DD & 68.2 $\pm$ 4.3 & 68.9 $\pm$ 4.0 & \textbf{74.9 $\pm$ 3.4} \\
\hline
NCI1 & 65.8 $\pm$ 2.7 & 71.5 $\pm$ 2.8 & \textbf{73.8 $\pm$ 2.0} \\
\hline
PROTEINS & 66.6 $\pm$ 3.2 & 66.0 $\pm$ 3.3 & \textbf{71.1 $\pm$ 3.1} \\
\hline
ENZYMES & 25.7 $\pm$ 6.0 & 29.5 $\pm$ 5.2 & \textbf{36.8 $\pm$ 5.8} \\
\hline
IMDB-B & 60.2 $\pm$ 4.3 & 64.3 $\pm$ 3.6 & \textbf{65.9 $\pm$ 2.2} \\
\hline
IMDB-M & 39.6 $\pm$ 3.4 & 35.3 $\pm$ 2.6 & \textbf{43.9 $\pm$ 2.4} \\
\hline
REDDIT-B & 78.0 $\pm$ 2.7 & 88.1 $\pm$ 2.1 & \textbf{89.6 $\pm$ 1.5} \\
\hline
REDDIT-5K & 46.4 $\pm$ 2.0 & 52.3 $\pm$ 1.5 & \textbf{52.8 $\pm$ 1.4} \\
\hline
COLLAB & 68.0 $\pm$ 2.3 & 71.0 $\pm$ 2.1 & \textbf{73.5 $\pm$ 2.2} \\
\hline
\end{tabular}
}
\end{table}

\noindent To verify the necessity of tuning LDP hyperparameters, we set the default values and vary a single hyperparameter at a time. We used 50 bins, histogram aggregation, normalization per graph, and linear scale. Due to space constraints, we do not include the whole results tables, but they are available on \href{https://github.com/j-adamczyk/LTP}{GitHub}. For each hyperparameter, we calculate the number of times each value gave the best result. Additionally, for each value, we also calculate the absolute average difference between its result and the best result for a given hyperparameter on each dataset. The lower the absolute difference, the better, since it means that a given hyperparameter value, on average, gives the best results among all its possible values. results are presented in Table \ref{table_ldp_hyperparameters}.

For the \textit{number of bins}, we can clearly select 50 bins as the optimal value. While 30 bins gave the best results the same number of times, on average they performed worse compared to the optimal hyperparameter value. Similarly, for \textit{normalization} it is evident that we do not need to perform any kind of normalization, since using no normalization obtained the best results on majority of datasets and on average. This is somewhat contrary to the results obtained in LDP paper \cite{cai2018simple}, but it is apparently an advantage of RF, since it considers each feature separately, while calculating tree splits. For \textit{aggregation} method, the results are very close, both for number of wins and average difference compared to the best result. In this case, the choice does not matter that much, and we choose the simpler histogram method. The linear \textit{scale} obtained much better results on average than the log scale, so the choice is obvious. Overall, this means that we can confidently recommend default values for all LDP hyperparameters, and tuning them is not particularly helpful. This dramatically decreases the computational cost, while having little effect on accuracy on average, which is a desirable tradeoff in a baseline method.

\begin{table}[htb]
\caption{Number of wins and absolute average difference between the result for a given hyperparameter value and the best result for any hyperparameter value. Higher number of wins is better, lower absolute average difference is better. For each hyperparameter, the value with the lowest absolute average difference has been marked in bold.}
\label{table_ldp_hyperparameters}
\scriptsize
\makebox[\textwidth][c]{
\begin{tabular}{|l|l|c|c|}
\hline
\textbf{Hyperparameter} & \textbf{Value} & \textbf{\# Wins} & \textbf{\begin{tabular}[c]{@{}l@{}}Abs. avg. difference\\ compared to best\end{tabular}} \\ \hline
\multirow{4}{*}{Number of bins} & 30              & 4          & 0.69\%          \\ \cline{2-4} 
                                & \textbf{50}     & \textbf{4} & \textbf{0.22\%} \\ \cline{2-4} 
                                & 70              & 1          & 0.70\%          \\ \cline{2-4} 
                                & 100             & 0          & 0.63\%          \\ \hline
\multirow{3}{*}{Normalization}  & \textbf{None}   & \textbf{5} & \textbf{0.23\%} \\ \cline{2-4} 
                                & Graph           & 2          & 2.03\%          \\ \cline{2-4} 
                                & Dataset         & 2          & 0.39\%          \\ \hline
\multirow{2}{*}{Aggregation}    & Histogram       & 5          & 0.71\%          \\ \cline{2-4} 
                                & \textbf{EDF}    & \textbf{4} & \textbf{0.69\%} \\ \hline
\multirow{2}{*}{Scale}          & \textbf{Linear} & \textbf{4} & \textbf{0.35\%} \\ \cline{2-4} 
                                & Log             & 5          & 0.64\%          \\ \hline
\end{tabular}
}
\end{table}

\noindent To assess whether additional descriptors increase accuracy of this method, we performed another set of experiments. We start with basic LDP, and add one additional descriptor at a time: lengths of shortest paths (SP), edge betweenness centrality (EBC), Jaccard Index (JI) and Local Degree Score (LDS). Finally, we check our proposed Local Topological Profile (LTP) method, combining LDP with EBC, JI and LDS descriptors. In all experiments, we keep the same hyperparameters: 50 bins, histogram aggregation, no normalization, and linear scale. As shown in Table \ref{table_additional_descriptors}, in every case the additional descriptors achieved the best result, while LTP was the best on 6 out of 9 datasets. On PROTEINS and IMDB-M it was the second best, being worse than the best by just $0.1\%$ on the latter. It was also the third best on NCI1. Therefore, we can conclude that adding selected descriptors definitely increases the discriminatory power of this method. LTP is a robust method, performing the best on average, and using it eliminates the need to tune descriptor selection.

\begin{figure}[htb]
\begin{center}
\includegraphics[width=0.8\textwidth]{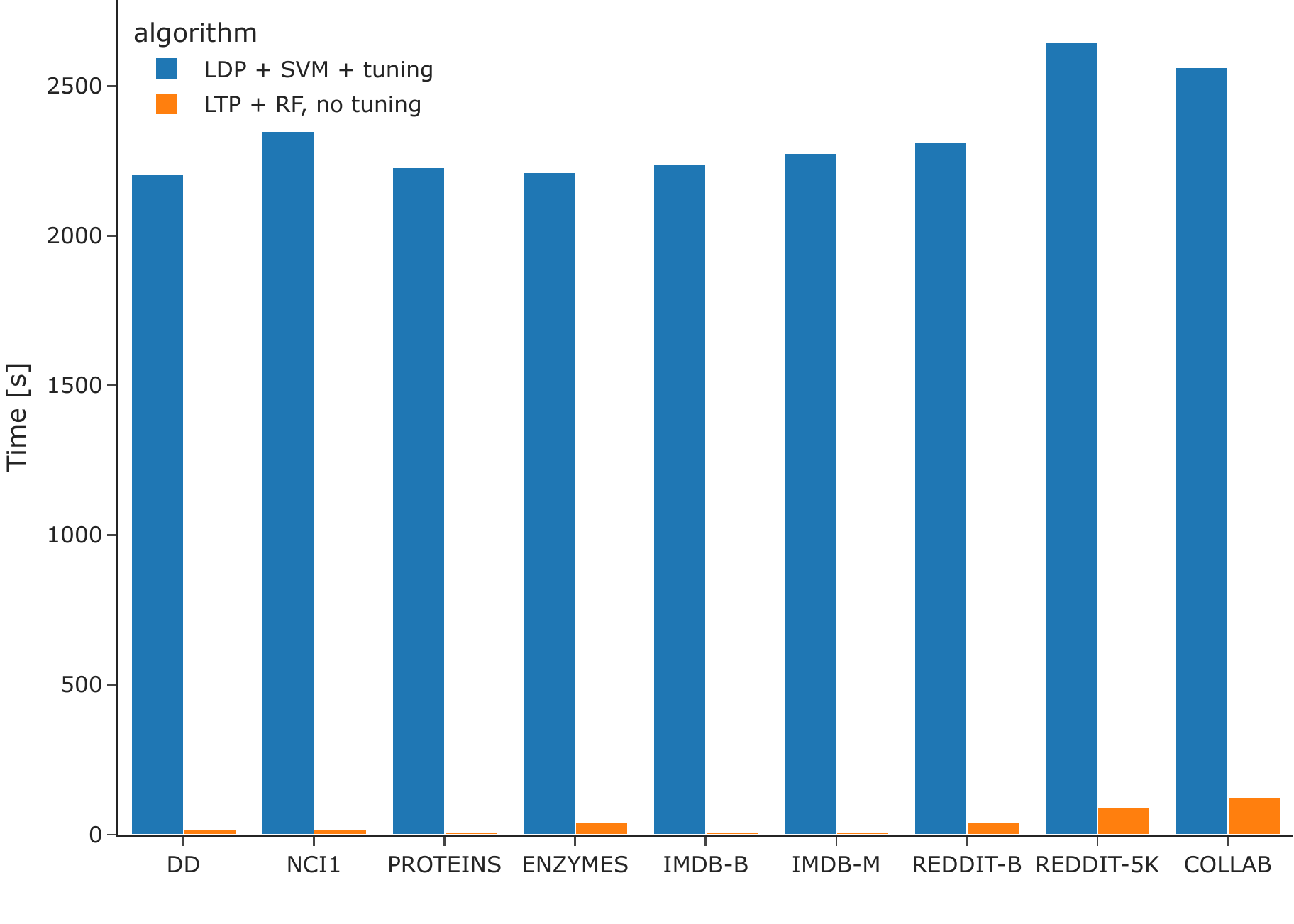}
\caption{Experiment time using LDP and LTP approaches.}    
\label{fig:ldp_vs_ltp}
\vspace{-0.5cm}
\end{center}
\end{figure}

For performance analysis, we compare the original method from \cite{cai2018simple}, i.e., LDP + SVM + hyperparameter tuning, with the proposed method, i.e. LTP + RF, without any tuning. For the former, we tune feature extraction hyperparameter and SVM hyperparameters separately, since grid search on combined parameters grids would result in approximately 20 times larger number of models to be trained for a single test fold, which is infeasible. We measure the time for the whole experimental procedure, i.e. feature extraction and classifier training and tuning for all 10 test folds. This way, we take all characteristics of both approaches into consideration: LDP taking more time due to tuning, and LTP due to extracting more features. As shown in Figure \ref{fig:ldp_vs_ltp}, our proposed LTP approach is vastly superior to LDP in terms of speed, being 1–3 orders of magnitude faster on all datasets. It should be noted that our LDP implementation is nevertheless much faster than the original one, since we use PyTorch Geometric to compute LDP features in parallel with optimized C++ subroutines. The original Python-based, sequential implementation in NetworkX would be additionally 1–2 orders of magnitude slower, based on preliminary experiments. Our method was also very fast on datasets with large number of large graphs (REDDIT datasets and COLLAB), which indicates good scalability. This is especially important considering that graph datasets are getting larger and baselines also have to scale well.

Lastly, we compare accuracy of LTP to GNNs from \cite{errica2019fair}, based on the same fair evaluation framework (compatible settings for model selection and model evaluation). For social networks, we compare against stronger models, using node degree. The  outcome is summarized in Table  \ref{table_comparison}. For easier comparison, in Table \ref{table_ranks} we also present the average rank of the model across all datasets, i.e. on average which place, from 1st to 8th, it took. Our LTP approach achieves state-of-the-art results on IMDB-B, IMDB-M, REDDIT-B and COLLAB, achieving as much as 3.8\% higher accuracy on COLLAB than the previous best method, GIN. Note that our method makes use of graph topology exclusively, ignoring node and edge features. This explains why on bioinformatics datasets we did not get as good results. In fact, on DD, PROTEINS and ENZYMES the best result is achieved by exclusively feature-based baseline from \cite{errica2019fair}, which does not use graph topology at all. On average, LTP obtained the best rank among all models, beating even a theoretically very powerful GIN architecture. Additionally, all GNNs require GPUs and many hours of computation, while our method gives results in mere seconds.

\begin{table}[htb]
\caption{Classification accuracy of LDP with additional descriptors and LTP. The best result for each dataset is marked in bold.}
\label{table_additional_descriptors}
\scriptsize
\centering
\begin{adjustbox}{max width=\textwidth}
\begin{tabular}{|l|c|c|c|c|c|c|}
\hline
\multicolumn{1}{|c|}{\textbf{Dataset}} &
  \textbf{LDP} &
  \textbf{\begin{tabular}[c]{@{}c@{}}LDP\\ + SP\end{tabular}} &
  \textbf{\begin{tabular}[c]{@{}c@{}}LDP\\ + EBC\end{tabular}} &
  \textbf{\begin{tabular}[c]{@{}c@{}}LDP\\ + JI\end{tabular}} &
  \textbf{\begin{tabular}[c]{@{}c@{}}LDP\\ + LDS\end{tabular}} &
  \textbf{LTP} \\ \hline
DD        & 76.0 $\pm$ 3.0 & 76.3 $\pm$ 2.8 & 77.0 $\pm$ 3.6          & 76.0 $\pm$ 3.4 & 75.8 $\pm$ 2.6          & \textbf{77.1 $\pm$ 3.7} \\ \hline
NCI1      & 77.2 $\pm$ 1.5 & 76.1 $\pm$ 1.6 & 76.8 $\pm$ 1.7          & 76.6 $\pm$ 1.4 & \textbf{77.4 $\pm$ 1.6} & 77.0 $\pm$ 1.9          \\ \hline
PROTEINS  & 70.6 $\pm$ 1.7 & 71.9 $\pm$ 2.2 & \textbf{73.0 $\pm$ 3.2} & 72.6 $\pm$ 3.2 & 71.4 $\pm$ 3.0          & 72.7 $\pm$ 4.2          \\ \hline
ENZYMES   & 37.4 $\pm$ 4.0 & 37.2 $\pm$ 5.4 & 40.2 $\pm$ 6.5          & 40.0 $\pm$ 6.6 & 38.7 $\pm$ 5.6          & \textbf{42.5 $\pm$ 4.1} \\ \hline
IMDB-B    & 71.3 $\pm$ 3.3 & 72.2 $\pm$ 4.0 & 72.9 $\pm$ 4.6          & 73.0 $\pm$ 4.3 & 74.2 $\pm$ 4.2          & \textbf{74.5 $\pm$ 4.3} \\ \hline
IMDB-M    & 49.0 $\pm$ 4.4 & 49.2 $\pm$ 4.1 & 49.2 $\pm$ 5.0          & 49.3 $\pm$ 4.5 & \textbf{50.1 $\pm$ 4.8} & 50.0 $\pm$ 4.6          \\ \hline
REDDIT-B  & 89.6 $\pm$ 1.2 & 90.5 $\pm$ 2.1 & 90.1 $\pm$ 1.7          & 89.6 $\pm$ 1.3 & 91.1 $\pm$ 1.1          & \textbf{91.1 $\pm$ 1.0} \\ \hline
REDDIT-5K & 51.9 $\pm$ 1.6 & 51.9 $\pm$ 1.9 & 51.7 $\pm$ 1.8          & 52.7 $\pm$ 2.0 & 53.1 $\pm$ 1.9          & \textbf{53.3 $\pm$ 1.5} \\ \hline
COLLAB    & 75.7 $\pm$ 2.0 & 76.5 $\pm$ 2.2 & 76.8 $\pm$ 1.9          & 76.8 $\pm$ 1.8 & 78.7 $\pm$ 2.4          & \textbf{79.4 $\pm$ 2.5} \\ \hline
\end{tabular}
\end{adjustbox}
\end{table}

\begin{table}[htb]
\caption{Comparison of accuracy with fair comparison results from \cite{errica2019fair}. Higher is better. Best result for each dataset has been marked in bold.}
\label{table_comparison}
\scriptsize
\centering

\begin{tabular}{|l|c|c|c|c|c|c|c|c|}
\hline
\multicolumn{1}{|c|}{\textbf{Dataset}} & \textbf{Baseline \cite{errica2019fair}}                                                & \textbf{DGCNN}                                          & \textbf{DiffPool}                                       & \textbf{ECC}                                            & \textbf{GIN}                                                     & \textbf{GraphSAGE}                                      & \textbf{LDP}                                            & \textbf{LTP}                                                     \\ \hline
DD                                     & \textbf{\begin{tabular}[c]{@{}c@{}}78.4\\ $\pm$4.5\end{tabular}} & \begin{tabular}[c]{@{}c@{}}76.6\\ $\pm$4.3\end{tabular} & \begin{tabular}[c]{@{}c@{}}75.0\\ $\pm$3.5\end{tabular} & \begin{tabular}[c]{@{}c@{}}72.6\\ $\pm$4.1\end{tabular} & \begin{tabular}[c]{@{}c@{}}75.3\\ $\pm$2.9\end{tabular}          & \begin{tabular}[c]{@{}c@{}}72.9\\ $\pm$2.0\end{tabular} & \begin{tabular}[c]{@{}c@{}}76.0\\ $\pm$3.0\end{tabular} & \begin{tabular}[c]{@{}c@{}}77.1\\ $\pm$3.7\end{tabular}          \\ \hline
NCI1                                   & \begin{tabular}[c]{@{}c@{}}69.8\\ $\pm$2.2\end{tabular}          & \begin{tabular}[c]{@{}c@{}}76.4\\ $\pm$1.7\end{tabular} & \begin{tabular}[c]{@{}c@{}}76.9\\ $\pm$1.9\end{tabular} & \begin{tabular}[c]{@{}c@{}}76.2\\ $\pm$1.4\end{tabular} & \textbf{\begin{tabular}[c]{@{}c@{}}80.0\\ $\pm$1.4\end{tabular}} & \begin{tabular}[c]{@{}c@{}}76.0\\ $\pm$1.8\end{tabular} & \begin{tabular}[c]{@{}c@{}}77.2\\ $\pm$1.5\end{tabular} & \begin{tabular}[c]{@{}c@{}}77.0\\ $\pm$ 1.9\end{tabular}         \\ \hline
PROTEINS                               & \textbf{\begin{tabular}[c]{@{}c@{}}75.8\\ $\pm$3.7\end{tabular}} & \begin{tabular}[c]{@{}c@{}}72.9\\ $\pm$3.5\end{tabular} & \begin{tabular}[c]{@{}c@{}}73.7\\ $\pm$3.5\end{tabular} & \begin{tabular}[c]{@{}c@{}}72.3\\ $\pm$3.4\end{tabular} & \begin{tabular}[c]{@{}c@{}}73.3\\ $\pm$4.0\end{tabular}          & \begin{tabular}[c]{@{}c@{}}73.0\\ $\pm$4.5\end{tabular} & \begin{tabular}[c]{@{}c@{}}70.6\\ $\pm$1.7\end{tabular} & \begin{tabular}[c]{@{}c@{}}72.7\\ $\pm$4.2\end{tabular}          \\ \hline
ENZYMES                                & \textbf{\begin{tabular}[c]{@{}c@{}}65.2\\ $\pm$6.4\end{tabular}} & \begin{tabular}[c]{@{}c@{}}38.9\\ $\pm$5.7\end{tabular} & \begin{tabular}[c]{@{}c@{}}59.5\\ $\pm$5.6\end{tabular} & \begin{tabular}[c]{@{}c@{}}29.5\\ $\pm$8.2\end{tabular} & \begin{tabular}[c]{@{}c@{}}59.6\\ $\pm$4.5\end{tabular}          & \begin{tabular}[c]{@{}c@{}}58.2\\ $\pm$6.0\end{tabular} & \begin{tabular}[c]{@{}c@{}}37.4\\ $\pm$4.0\end{tabular} & \begin{tabular}[c]{@{}c@{}}42.5\\ $\pm$4.1\end{tabular}          \\ \hline
IMDB-B                                 & \begin{tabular}[c]{@{}c@{}}70.8\\ $\pm$5.0\end{tabular}          & \begin{tabular}[c]{@{}c@{}}69.2\\ $\pm$3.0\end{tabular} & \begin{tabular}[c]{@{}c@{}}68.4\\ $\pm$3.3\end{tabular} & \begin{tabular}[c]{@{}c@{}}67.7\\ $\pm$2.8\end{tabular} & \begin{tabular}[c]{@{}c@{}}71.2\\ $\pm$3.9\end{tabular}          & \begin{tabular}[c]{@{}c@{}}68.8\\ $\pm$4.5\end{tabular} & \begin{tabular}[c]{@{}c@{}}71.3\\ $\pm$3.3\end{tabular} & \textbf{\begin{tabular}[c]{@{}c@{}}74.5\\ $\pm$4.3\end{tabular}} \\ \hline
IMDB-M                                 & \begin{tabular}[c]{@{}c@{}}49.1\\ $\pm$3.5\end{tabular}          & \begin{tabular}[c]{@{}c@{}}45.6\\ $\pm$3.4\end{tabular} & \begin{tabular}[c]{@{}c@{}}45.6\\ $\pm$3.4\end{tabular} & \begin{tabular}[c]{@{}c@{}}43.5\\ $\pm$3.1\end{tabular} & \begin{tabular}[c]{@{}c@{}}48.5\\ $\pm$3.3\end{tabular}          & \begin{tabular}[c]{@{}c@{}}47.6\\ $\pm$3.5\end{tabular} & \begin{tabular}[c]{@{}c@{}}49.0\\ $\pm$4.4\end{tabular} & \textbf{\begin{tabular}[c]{@{}c@{}}50.0\\ $\pm$4.6\end{tabular}} \\ \hline
REDDIT-B                               & \begin{tabular}[c]{@{}c@{}}82.2\\ $\pm$3.0\end{tabular}          & \begin{tabular}[c]{@{}c@{}}87.8\\ $\pm$2.5\end{tabular} & \begin{tabular}[c]{@{}c@{}}89.1\\ $\pm$1.6\end{tabular} & OOR                                                     & \begin{tabular}[c]{@{}c@{}}89.9\\ $\pm$1.9\end{tabular}          & \begin{tabular}[c]{@{}c@{}}84.3\\ $\pm$1.9\end{tabular} & \begin{tabular}[c]{@{}c@{}}89.6\\ $\pm$1.2\end{tabular} & \textbf{\begin{tabular}[c]{@{}c@{}}91.1\\ $\pm$1.0\end{tabular}} \\ \hline
REDDIT-5K                              & \begin{tabular}[c]{@{}c@{}}52.2\\ $\pm$1.5\end{tabular}          & \begin{tabular}[c]{@{}c@{}}49.2\\ $\pm$1.2\end{tabular} & \begin{tabular}[c]{@{}c@{}}53.8\\ $\pm$1.4\end{tabular} & OOR                                                     & \textbf{\begin{tabular}[c]{@{}c@{}}56.1\\ $\pm$1.7\end{tabular}} & \begin{tabular}[c]{@{}c@{}}50.0\\ $\pm$1.3\end{tabular} & \begin{tabular}[c]{@{}c@{}}51.9\\ $\pm$1.6\end{tabular} & \begin{tabular}[c]{@{}c@{}}53.3\\ $\pm$1.5\end{tabular}          \\ \hline
COLLAB                                 & \begin{tabular}[c]{@{}c@{}}70.2\\ $\pm$1.5\end{tabular}          & \begin{tabular}[c]{@{}c@{}}71.2\\ $\pm$1.9\end{tabular} & \begin{tabular}[c]{@{}c@{}}68.9\\ $\pm$2.0\end{tabular} & OOR                                                     & \begin{tabular}[c]{@{}c@{}}75.6\\ $\pm$2.3\end{tabular}          & \begin{tabular}[c]{@{}c@{}}73.9\\ $\pm$1.7\end{tabular} & \begin{tabular}[c]{@{}c@{}}75.7\\ $\pm$2.0\end{tabular} & \textbf{\begin{tabular}[c]{@{}c@{}}79.4\\ $\pm$2.5\end{tabular}} \\ \hline
\end{tabular}
\end{table}

\begin{table}[!htb]
\caption{Comparison of average model ranks. The best result is marked in bold.}
\label{table_ranks}
\scriptsize
\centering
\begin{tabular}{|c|c|c|c|c|c|c|c|c|}
\hline
\textbf{} &
  \textbf{Baseline \cite{errica2019fair}} &
  \textbf{DGCNN} &
  \textbf{DiffPool} &
  \textbf{ECC} &
  \textbf{GIN} &
  \textbf{GraphSAGE} &
  \textbf{LDP} &
  \textbf{LTP} \\ \hline
\textbf{\begin{tabular}[c]{@{}c@{}}Average\\ rank\end{tabular}} &
  3.8 &
  5.2 &
  4.6 &
  7.6 &
  2.7 &
  5.4 &
  4 &
  \textbf{2.6} \\ \hline
\end{tabular}
\end{table}

\section{Conclusions}
\label{conclusions}

We presented the new structural baseline for graph classification called Local Topological Profile (LTP). The research questions addressing its efficiency and scalability in comparison to related LDP method and competitive GNN methods were studied in the experimental section, where we conclude that using the Random Forest classifier instead of SVM improved the accuracy and the speed of computation by a large margin and this observation applies to all datasets used. We note that tuning of feature extraction hyperparameters is not necessary therefore, we can use default values for all datasets, decreasing tuning time significantly. More importantly, we observe that introducing additional topological descriptors increases predictive accuracy in LTP method significantly. Using all three proposed descriptors (Edge Betweenness Centrality, Jaccard Index, Local Degree Score) gives very good prediction results across nine benchmark datasets, at the same time LTP is 2–3 orders of magnitude faster than the original LDP approach. Finally, we achieve state-of-the-art results on 4 out of 9 benchmark datasets, and in other cases get very strong accuracy, comparable to or even outcompeting modern GNNs, while using exclusively the graph topology. We share the software package with the research community, hoping that it can be useful in comparing results achieved by state-of-the-art graph classification models.

In our future work, we plan to extend the number of vertex/edge descriptors and enrich the expressive power of LTP towards more global features such as \textit{eccentricity}. We also plan to merge LTP feature extraction and baselines from \cite{errica2019fair}, to strengthen performance on more feature-focused bioinformatics datasets.

\subsubsection{Acknowledgements} The research presented in this paper was financed from the funds assigned by Polish Ministry of Science and Higher Education to AGH University of Science and Technology. We would like to thank Alexandra Elbakyan for her work and support for accessibility of science.

\bibliographystyle{splncs04}
\bibliography{mybibliography}

\end{document}